\documentclass[10pt,twocolumn,letterpaper]{article}

\usepackage[pagenumbers]{cvpr}
\usepackage{placeins}

\usepackage{cuted} %< for strip environment

\usepackage{currfile} %< for \currfilebase macro

\usepackage{caption} %< for "\captionof" commands (teaser)

\usepackage{booktabs}
\usepackage{array}
\usepackage{float}
\usepackage{amssymb}
\usepackage{pifont}

\definecolor{cvprblue}{rgb}{0.21,0.49,0.74}
\usepackage[breaklinks,colorlinks,allcolors=cvprblue]{hyperref}

\title{PGS-PCQA: Perceptually Gated Synthetic Supervision for No-Reference Point Cloud Quality Assessment}

\author{Qinyuan Fan\textsuperscript{1,}\thanks{Corresponding author: \href{mailto:qinyuan.fan@tu-berlin.de}{qinyuan.fan@tu-berlin.de}} \quad Meng Zhang\textsuperscript{2} \quad Clemens G\"{u}hmann\textsuperscript{1}\\
{\normalsize \textsuperscript{1}Chair of Electronic Measurement and Diagnostic Technology, Technische Universit\"{a}t Berlin, Berlin 10587, Germany}\\
{\normalsize \textsuperscript{2}Department of Mechanical Engineering, Politecnico di Milano}}

\begin{document}
\maketitle

\begin{abstract}
Existing no-reference point cloud quality assessment (NR-PCQA) methods perform well within individual datasets but generalize poorly across datasets. We identify insufficient distortion coverage in training data as a key cause: on unseen distortion types, models rank severity levels near chance or even in reverse. We propose perceptually gated synthetic supervision for no-reference point cloud quality assessment (PGS-PCQA), an NR-PCQA model trained with synthetic distortion ranking. It builds a synthetic distortion bank from source reference point clouds and uses within-type severity ranking as supervision in addition to the source mean opinion scores (MOS), requiring no subjective scores for the synthetic samples. Since higher severity does not always mean lower perceived quality, a perceptual gate softens the ranking targets of pairs with small perceptual differences in rendered views. The synthetic supervision is used only during training and needs no target-domain data. In a single training run on SJTU-PCQA, PGS-PCQA achieves Spearman rank correlation coefficients (SRCCs) of 0.709 on WPC and 0.751 on BASICS. The corresponding published DSP-PCQA scores are 0.595 and 0.723. Under reference-disjoint cross-validation, PGS-PCQA achieves SRCCs of 0.955 on SJTU-PCQA, 0.934 on WPC, 0.906 on BASICS, and 0.791 on LS-PCQA Part I. In the exclusion experiment, the reported scores for the reduced synthetic bank represent 58\%--65\% of the full-bank SRCC gain relative to training without the bank. Code is available at \mbox{\href{https://github.com/Fan-QY/PGS-PCQA}{https://github.com/Fan-QY/PGS-PCQA}}.

\end{abstract}

\section{Introduction}
\label{sec:intro}

\begin{figure}[t]
  \centering
  \includegraphics[width=\linewidth]{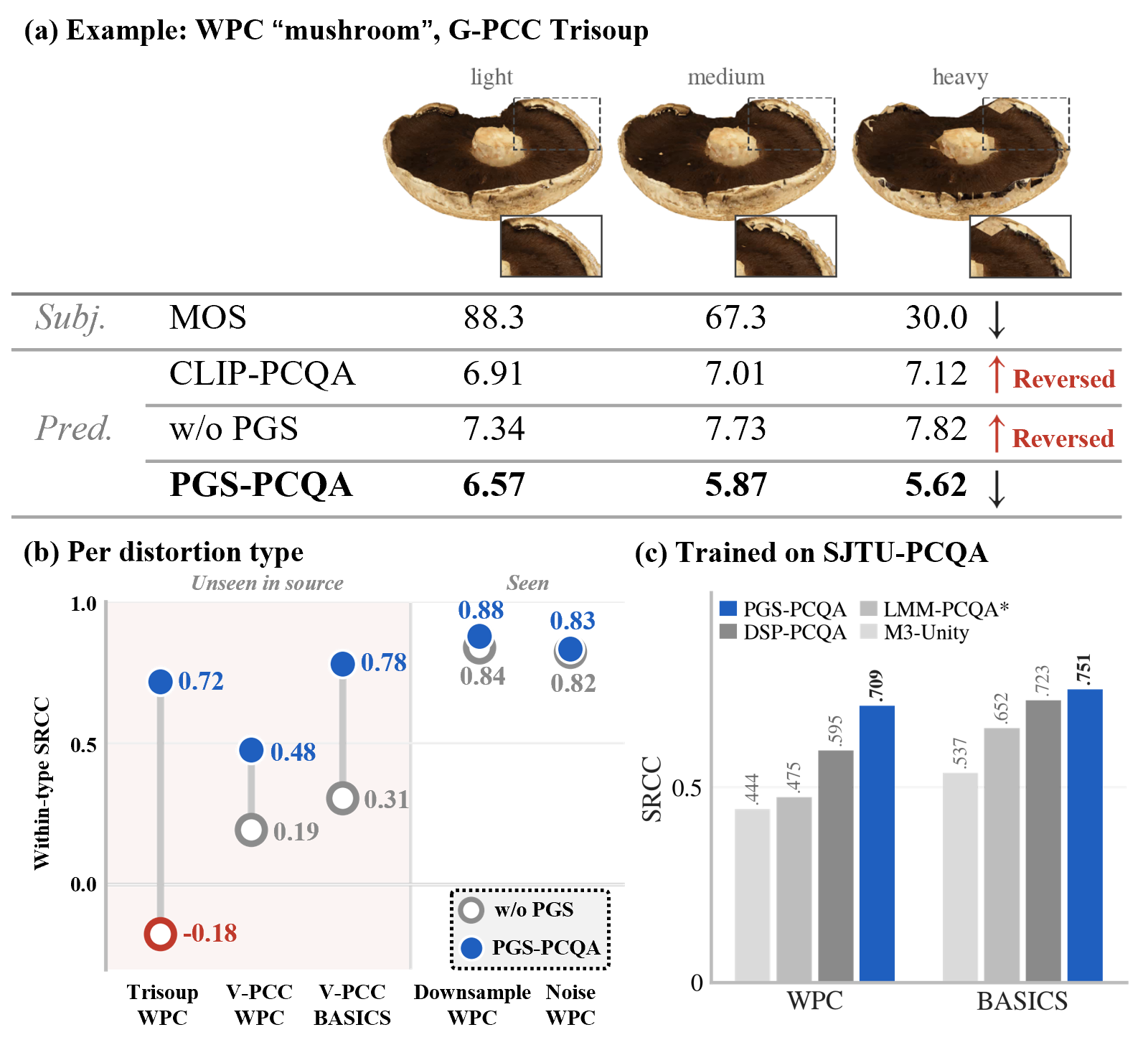}
  \caption{Models trained on SJTU-PCQA fail on distortion types absent from the source. (a) WPC mushroom under G-PCC Trisoup: MOS (0--100) drops with severity, while CLIP-PCQA and our model w/o PGS predict higher scores. PGS-PCQA follows the MOS. (b) Within-type SRCC. (c) Cross-dataset SRCC. The PGS-PCQA scores are from the run reported in \cref{tab:cross}. An asterisk marks baselines reproduced in DSP-PCQA.}
  \label{fig:teaser}
\end{figure}

Point cloud quality assessment (PCQA) predicts how geometric and color distortions affect the subjective quality of 3D content. Since the pristine reference is usually unavailable in practice, no-reference methods (NR-PCQA) are more practical~\cite{shan2024copa}. NR-PCQA has moved from local geometric features to multi-view images, point--image fusion, and vision--language models~\cite{shan2024copa,zhou2024structure,liu2025clippcqa,li2026pmp}; within a single dataset the %
SRCC now commonly exceeds 0.9, but it drops markedly on another dataset~\cite{shan2024copa,li2026dst,li2026dsp}. One reason is that the usual content-wise cross-validation lets training and test sets share the same distortion types, whereas real distortions come from different codecs, acquisition conditions, and reconstruction algorithms. Intra-dataset scores therefore do not show whether a model can handle unseen distortions.

Existing improvements follow two directions: learning better quality features or narrowing the domain gap. CoPA~\cite{shan2024copa} learns point cloud representations with quality-aware contrastive pre-training, CLIP-PCQA~\cite{liu2025clippcqa} introduces a pretrained vision--language model, and PMP-PCQA~\cite{li2026pmp} and DSP-PCQA~\cite{li2026dsp} strengthen the modeling of geometric, visual, and semantic information; DST-PCQA~\cite{li2026dst} uses target-domain data to select relevant synthetic distortions. We study a different question: without any target-domain data, using only the source dataset, can a model judge distortions it has never seen?

We first break the cross-dataset results down by distortion type. Trained on SJTU-PCQA and tested on WPC, models rank the downsampling and noise found in the source dataset correctly, but rank the coding distortions absent from it at chance level or even in reverse: on the Trisoup mode of geometry-based point cloud compression (G-PCC), the within-type SRCC is about 0 for the original CLIP-PCQA and $-0.18$ for our model with its perceptually gated synthetic supervision (PGS) removed (\cref{fig:teaser}). When the training data lack a distortion type, the model has no basis for ordering its quality.

A direct remedy is to generate distortions ourselves: applying compression, geometric, and color degradations to the source references yields many distortion types and levels without new content. These samples have no subjective scores and cannot be used for regression, but the level order within a type can serve as relative supervision, as in image quality assessment~\cite{liu2017rankiq} and point cloud geometry quality assessment~\cite{su2022prlgqa,li2025lrlgqa}. We use this level order to extend the training supervision of NR-PCQA, without generating pseudo mean opinion scores (MOS) for synthetic samples.

The catch is that a higher level does not always look worse. Some parameter changes are barely visible in the renders, and some distortions change appearance inconsistently with their nominal severity. On three real datasets, subjective scores follow the preset order for only 0.80--0.87 of the 20\% of level pairs with the smallest rendered-distance difference (\cref{tab:gatemos}). Treating every level pair as a definite quality order forces the model to separate samples that look the same. dipIQ~\cite{ma2017dipiq} estimates the reliability of image pairs with full-reference metrics and down-weights uncertain pairs. Synthetic point clouds come with their references, so we can directly measure how far each distorted version deviates from the reference in rendered views~\cite{zhang2018lpips} and judge whether a level pair has visual support. This distance is not a subjective score, so we use it only to set how strong the ranking target is, not to create quality labels.

We therefore propose PGS-PCQA. It builds a synthetic distortion bank from the source references (G-PCC, video-based point cloud compression (V-PCC)~\cite{schwarz2019mpeg,graziosi2020overview}, and common geometric and color degradations; 84 versions per reference), pairs different levels of the same reference and distortion type, and trains with these pairs together with MOS supervision on real samples. A perceptual gate sets a soft ranking target from the difference of the two samples' rendered distances to the reference: when the difference is clear and agrees with the level order, the lighter level must score higher; when it is small or reversed, the two scores are only pulled together without imposing an order. The synthetic data and references are used only for training; at inference the model remains no-reference and needs no target-domain data.

We evaluate PGS-PCQA on SJTU-PCQA~\cite{yang2021sjtu}, WPC~\cite{liu2023wpc}, BASICS~\cite{ak2024basics}, and LS-PCQA~\cite{liu2023lspcqa}, with per-distortion-type analysis, supervision ablations, and target-type exclusion experiments. Our contributions are:
\begin{enumerate}
  \item PGS-PCQA, a model that supplements quality supervision with the level order of source-domain synthetic distortions and relaxes visually unreliable level pairs with a perceptual gate, requiring no extra subjective scores.
  \item A per-distortion-type analysis of cross-dataset results, showing that models rank coding distortions unseen in training at chance level or even in reverse.
  \item On four public datasets, we evaluate PGS-PCQA under
intra-dataset and cross-dataset protocols. On SJTU-PCQA$\to$WPC,
PGS-PCQA achieves an SRCC of 0.709, compared with 0.595 for
DSP-PCQA. The exclusion experiment also examines transfer to
distortion categories removed from the synthetic bank.
\end{enumerate}

\section{Related Work}
\label{sec:related}

\subsection{PCQA}

\noindent\textbf{Full-reference methods}
compare a distorted point cloud with its reference, including point-to-point~\cite{mekuria2017codec} and point-to-plane~\cite{tian2017geometric} errors and the point cloud quality metric PCQM~\cite{meynet2020pcqm}, the graph-signal-based similarity GraphSIM~\cite{yang2022graphsim}, and the transformational-complexity-based distortion metric TCDM~\cite{zhang2024tcdm}, which exploit curvature, color, or structural cues. They require the reference at test time, whereas we use references only during training and only for synthetic samples.

\noindent\textbf{No-reference methods}
do not require a reference. Point-based methods extract quality features directly from 3D geometry and color~\cite{zhang2022nss,liu2023lspcqa,shan2024gpanet,liu2026egms}; projection-based methods render multi-view images and model visual distortions with image networks~\cite{yang2022itpcqa,zhang2024gms3dqa,mu2024hallucinated,xiong2026safnet,seo2026r3pcqa}; multi-modal methods combine point and image features~\cite{zhang2023mmpcqa,zhou2024m3unity,wu2025danet,li2026pmp,li2026dsp,su2026compcqa}. Other works improve quality learning with contrastive pre-training~\cite{shan2024copa}, coding quantization parameters~\cite{duan2025dqp}, or the stochasticity of subjective ratings~\cite{fan2025stochasticity}. CLIP-PCQA~\cite{liu2025clippcqa} brings contrastive language--image pre-training (CLIP)~\cite{radford2021clip} into NR-PCQA and predicts a distribution over quality levels with learnable prompts; LMM-PCQA~\cite{zhang2024lmmpcqa} and Dynamic Alignment~\cite{wu2026dynamic} further introduce large multimodal models. These methods mainly improve the quality representation, while supervision still comes entirely from the subjective scores of the dataset, so distortion types absent from the training data receive no quality supervision. Built on CLIP-PCQA, PGS-PCQA supplies this missing information through its training supervision.

\begin{figure*}[t]
  \centering
  \includegraphics[width=\textwidth]{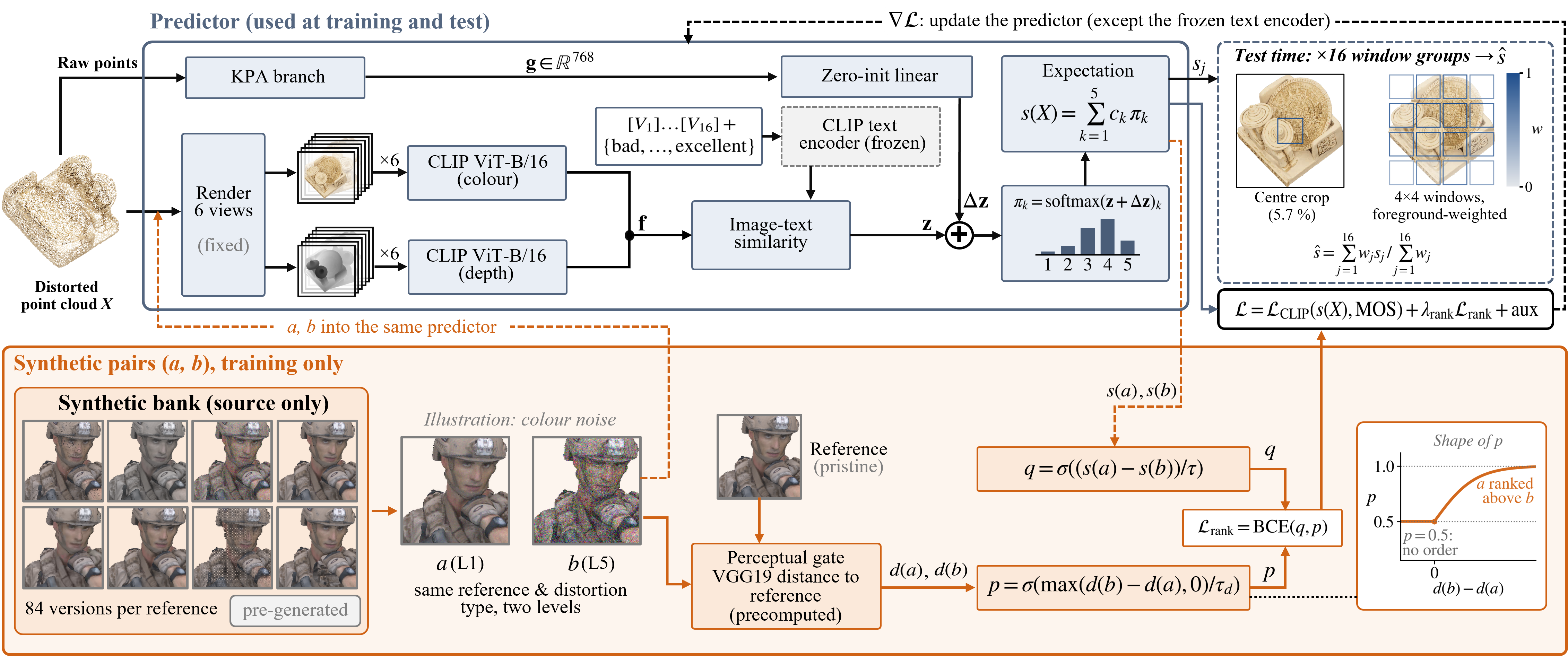}
  \caption{Overview of PGS-PCQA. Top: the no-reference predictor used at inference, with foreground-weighted $4\times4$ testing at the top right (\cref{sec:tta}). Bottom (training only): a synthetic bank pre-generated from source references provides pairs $(a,b)$ of the same reference and distortion type at different levels; their VGG19 rendered distances to the reference set, via the perceptual gate, the soft target $p$ of $\mathcal{L}_{\text{rank}}$. Real samples are supervised by MOS ($\mathcal{L}_{\text{CLIP}}$); the auxiliary and KPA heads are removed at inference.}
  \label{fig:overview}
\end{figure*}

\subsection{Synthetic Distortion and Ranking Supervision}

RankIQA~\cite{liu2017rankiq} orders same-type samples by their distortion parameters for ranking pre-training, and PRL-GQA~\cite{su2022prlgqa} and LRL-GQA~\cite{li2025lrlgqa} apply pairwise and listwise ranking to point cloud geometry quality assessment; CONTRIQUE, Re-IQA, and ARNIQA~\cite{madhusudana2022contrique,saha2023reiqa,agnolucci2024arniqa} use synthetic distortions as self-supervision for image quality representations; our pairwise loss follows the probabilistic form of RankNet~\cite{burges2005ranknet}. Automatically constructed orders are not always reliable: dipIQ~\cite{ma2017dipiq} measures the confidence of image pairs with several full-reference metrics and down-weights uncertain pairs. DST-PCQA~\cite{li2026dst} selects relevant synthetic distortions with unlabeled target data, and QD-PCQA~\cite{zhang2026qdpcqa} transfers image quality knowledge to unlabeled point clouds by quality-aware unsupervised domain adaptation; both require target-domain data. PGS-PCQA uses no target-domain data; it sets the ranking target of each level pair itself, rather than its loss weight, from a deep-feature distance on rendered views~\cite{zhang2018lpips,simonyan2015vgg}, and builds rankings only within the same reference and distortion type. The difference is that confidence weighting still asks the model to learn a definite order, only with less influence for some pairs, whereas the soft target states directly that two levels are hard to tell apart: with similar perceptual distances, the two scores are only pulled together without imposing an order. The gate only relaxes the level order and never reverses it, since perceptual distance is not a subjective score.

\section{Method}
\label{sec:method}

\subsection{Quality Prediction Framework}

\cref{fig:overview} shows the overall architecture. PGS-PCQA uses CLIP-PCQA~\cite{liu2025clippcqa} as the base quality predictor and extends its training supervision with synthetic distortion ranking. An input point cloud $X$ is rendered into color and depth images from six views; CLIP visual encoders~\cite{radford2021clip,dosovitskiy2021vit} extract image features $\mathbf{f}$, whose similarity to the text features of five quality levels gives level logits $\mathbf{z}$. The text prompts follow CoOp~\cite{zhou2022coop}: 16 learnable context vectors shared across levels are prepended to the fixed level words (bad, poor, fair, good, excellent) and encoded by the frozen CLIP text encoder; the context vectors and both CLIP image encoders are trained. The level probabilities are $\pi_k=\mathrm{softmax}(\mathbf{z}+\Delta\mathbf{z})_k$ and the quality score is their expectation $s(X)=\sum_{k=1}^{5}c_k\,\pi_k$, where $c_k$ is the anchor of level $k$ on the dataset's MOS scale (e.g., 2, 4, \ldots, 10 for SJTU-PCQA); higher is better, where $\Delta\mathbf{z}$ comes from the kernel point attention (KPA) geometry branch below.

On top of the base network we add two modules. The KPA geometry branch extracts a feature $\mathbf{g}\in\mathbb{R}^{768}$ from multi-scale neighborhoods of the raw points, maps it through a zero-initialized linear layer to the correction $\Delta\mathbf{z}$, and has its own MOS regression head. The distortion auxiliary head performs distortion-type classification and parameter regression on $\mathbf{f}$, with labels parsed from sample file names, and is removed at inference. The network is trained with the subjective scores of the source dataset, and the auxiliary head additionally with the distortion labels. Our core addition is synthetic distortion supervision on top of MOS supervision, requiring no subjective scores for the synthetic samples and letting the model learn degradations insufficiently covered by the source dataset. PGS is always measured against the same network without PGS.

\subsection{Synthetic Distortion Ranking}

For each source reference point cloud we generate 84 synthetic versions: 19 G-PCC-coded, 5 V-PCC-coded, and 60 generic geometric and color distortions. They cover different mechanisms and severities and are fixed before training without any target-domain data (composition in \cref{tab:bank}).

Synthetic samples have no subjective scores, but their generation parameters give distortion levels. We pick two samples of different levels from the same reference and distortion type, denoted $a$ (lighter) and $b$ (heavier). The predicted probability that $a$ has higher quality than $b$ is
\begin{equation}
  q = \sigma\big((s(a) - s(b))/\tau\big),
\end{equation}
where $\sigma$ is the sigmoid function and the temperature $\tau$, set to 5\% of the score range, controls how score differences map to probabilities.

\subsection{Perceptually Gated Ranking}\label{sec:gate}

To check whether a level order has visual support, we compute the distance between each synthetic sample and its reference in rendered views with multi-layer VGG19~\cite{simonyan2015vgg} features~\cite{zhang2018lpips}, averaged over six views. The supervision strength of each pair is thus set by the actual visual change in the renders, not by the levels alone. Pairs are built only within a distortion type; qualities of different types are not compared.

Let $d(a)$ and $d(b)$ be the perceptual distances of the two samples to their common reference. A larger distance for the heavier level $b$ supports the nominal order; close or reversed distances make it less reliable. We therefore build a soft ranking target from the distance difference:
\begin{equation}
  p = \sigma\big(\max(d(b) - d(a), 0)/\tau_d\big),
\end{equation}
where $\tau_d$, set to 0.1 times the standard deviation of synthetic perceptual distances, controls the influence of the difference. In the banks built from the four datasets, about 22\% of adjacent-level pairs receive a soft target below 0.75, and 3.5\%--4.9\% have a perceptual order opposite to the level order, mostly from color-only G-PCC, local shift and rotation, and saturation changes (\cref{tab:gate}). On the real subjective data of SJTU-PCQA, WPC, and BASICS, the Spearman correlation between the perceptual-distance difference and the MOS difference of level pairs is 0.78--0.85, higher than for the level difference itself (0.59--0.66); for adjacent levels only it is still 0.46--0.73, and subjective scores follow the preset order for 0.71--0.83 of the 20\% of adjacent pairs with the smallest $\Delta d$, versus 0.94--1.00 for the largest 20\% (\cref{tab:gatemos}). The pairs the gate relaxes most are those on which human judgments disagree most. A larger difference requires $s(a)$ to exceed $s(b)$ more clearly; when $d(b)\le d(a)$, $p=0.5$: no order is required and the two scores are only pulled toward each other. Hence $p\in[0.5,1)$: the gate only modulates the strength of the ranking constraint and never reverses the preset level order. The predicted probability $q$ is optimized against $p$ with binary cross-entropy (BCE)~\cite{burges2005ranknet}:
\begin{equation}
  \mathcal{L}_{\text{rank}} = \mathrm{BCE}(q, p).
\end{equation}

\subsection{Joint Optimization}

Real samples with subjective scores and synthetic pairs update the network jointly. The total loss is
\begin{equation}
  \mathcal{L} = \mathcal{L}_{\text{CLIP}} + \lambda_{\text{aux}}\mathcal{L}_{\text{aux}} + \lambda_{\text{KPA}}\mathcal{L}_{\text{KPA}} + \lambda_{\text{rank}}\mathcal{L}_{\text{rank}},
  \label{eq:total}
\end{equation}
where $\mathcal{L}_{\text{CLIP}}$ is the original subjective-quality loss of CLIP-PCQA, $\mathcal{L}_{\text{aux}}$ and $\mathcal{L}_{\text{KPA}}$ supervise the auxiliary task and the geometry branch, $\mathcal{L}_{\text{rank}}$ is the proposed perceptually gated ranking loss, and all weights $\lambda$ are 0.5.

The synthetic distortions and their ranking supervision are used only during training and add no inference cost. Implementation and ablations of the KPA branch and the auxiliary head are in \cref{sec:supp_1,sec:supp_2}: their gains vary across datasets, and removing KPA leaves the cross-dataset gain essentially unchanged (\cref{tab:ablation}).

\subsection{Foreground-Weighted Multi-Region Testing}
\label{sec:tta}

The CLIP encoders take $224\times224$ inputs, and the common center crop covers only a small part of each projected view (median area ratio 9\%–17\% across the four datasets), so distortions elsewhere never reach the predictor.
At test time we place $224\times224$ windows at the $4\times4$ uniform grid positions of every view; windows at the same relative position across the six views form one group, giving 16 groups. Group $j$ yields a score $s_j$ and a weight $w_j$, the foreground-pixel ratio of its windows averaged over the six views (top right of \cref{fig:overview}). The final score is
\begin{equation}
  \hat{s}=\sum_{j=1}^{16} w_j s_j \Big/ \sum_{j=1}^{16} w_j ,
\end{equation}
so windows on empty background receive almost no weight. GMS-3DQA~\cite{zhang2024gms3dqa} also samples grid patches over projections to cover the whole image, but it stitches them into a new input image on which the model is trained; our multi-region testing is used only at inference, leaves training unchanged, and weights windows by foreground ratio, so it applies directly to existing models (Com-PCQA in \cref{tab:ablation}(b)). The weights of the model are unchanged; inference time grows from about 0.15\,s to about 1.2\,s per sample (one RTX 2080 Ti, averaged over the 148 test samples of one WPC fold processed one by one, including image loading and KPA neighborhood computation).

\section{Experiments}
\label{sec:exp}

\subsection{Setup}

\noindent\textbf{Datasets}
used for evaluation are four datasets with MOS: SJTU-PCQA~\cite{yang2021sjtu} (378 distorted samples), WPC~\cite{liu2023wpc} (740), BASICS~\cite{ak2024basics} (1,494), and LS-PCQA~\cite{liu2023lspcqa} Part I (930 samples with real MOS), used in both intra-dataset and cross-dataset settings.

\noindent\textbf{Intra-dataset protocol.}
For intra-dataset evaluation, we group all distorted samples by their reference
point cloud and assign each reference to a single fold. We use nine folds for
SJTU-PCQA and five folds for WPC, BASICS, and LS-PCQA Part I. In each fold, we train
once for 50 epochs on the training references and evaluate the checkpoint at epoch
50 on the held-out references.

\noindent\textbf{Cross-dataset protocol.}
\Cref{tab:cross} reports results from one training run per source dataset.
Within each reported run, we select the checkpoint with the lowest
source-domain MOS training loss, excluding the ranking and auxiliary
terms, and use that checkpoint for all corresponding target datasets.
For transfers between SJTU-PCQA and LS-PCQA, we remove the shared
references from LS-PCQA before training when LS-PCQA is the source
and before evaluation when it is the target. When SJTU-PCQA is the
target, we evaluate all its samples.

The full-model results in \cref{tab:ablation}(a) and
\cref{tab:lto,tab:mech} use the SJTU-PCQA checkpoint reported in
\cref{tab:cross}. The remaining configurations in these ablation
tables use the mean SRCC from their available runs. The results in
\cref{tab:pertype,tab:lto_type,tab:mm_pgs,tab:calib} also summarize
their respective runs using mean SRCC.

\noindent\textbf{Metrics.}
SRCC is our primary metric and is computed from the raw predictions. For PLCC and
RMSE, we map the predictions to the dataset's MOS scale using a five-parameter
logistic fit on each test fold~\cite{vqeg2000,sheikh2006statistical}.
\Cref{tab:intra} reports the arithmetic mean of each metric across the folds. We
compare RMSE within each dataset because the MOS scales differ (1--10 for SJTU-PCQA,
0--100 for WPC, and 1--5 for BASICS and LS-PCQA).

\noindent\textbf{Implementation details}
are as follows: we train with Adam for 50 epochs with batch size 4, a learning rate of $4\times10^{-6}$ for the CLIP encoders and $5\times10^{-4}$ for the KPA branch, and two synthetic pairs per step. The model has about 175M trainable parameters. All hyperparameters are the same for all datasets and source--target pairs. Each intra-dataset fold trains on one RTX A5000 in about 1.9\,h (SJTU-PCQA) to 8\,h (BASICS, including the synthetic bank). The synthetic bank and perceptual distances are generated offline; details are in \cref{sec:supp_5}.

\subsection{Quality Prediction Results}

\begin{figure*}[t]
  \centering
  \includegraphics[width=\textwidth]{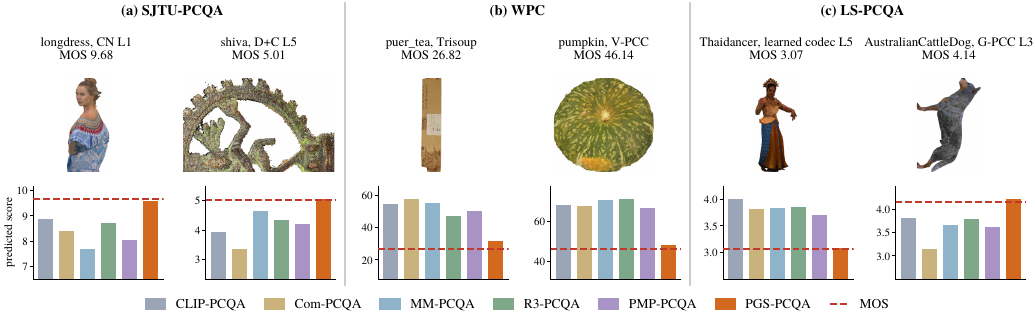}
  \caption{Qualitative comparison on SJTU-PCQA, WPC, and LS-PCQA Part I.
  Bars show held-out predictions mapped to the MOS scale by a per-fold
  logistic fit. The red dashed line shows the subjective MOS. The methods
  reproduced for this figure use the same reference folds. We reimplement
  PMP-PCQA and use the official code for the other comparison methods.
  \Cref{fig:qual_full} shows additional examples.}
  \label{fig:qual}
\end{figure*}

\noindent\textbf{Intra-dataset.}
\Cref{tab:intra} reports intra-dataset results under reference-disjoint
cross-validation with a fixed 50-epoch training budget. PGS-PCQA achieves
SRCCs of 0.955 on SJTU-PCQA, 0.934 on WPC, 0.906 on BASICS, and 0.791 on
LS-PCQA Part I. \Cref{fig:qual} shows predictions for held-out references,
and \cref{sec:supp_4} gives the full comparison and documents the baseline
protocols.

\begin{table*}[t]
  \centering
  \caption{Intra-dataset quality prediction results (SRCC$\uparrow$, PLCC$\uparrow$,
  RMSE$\downarrow$). PGS-PCQA is evaluated using reference-disjoint cross-validation
  with nine folds on SJTU-PCQA and five folds on WPC, BASICS, and LS-PCQA Part I.
  Each fold contributes one training run. We use the checkpoint at epoch 50 with
  foreground-weighted $4\times4$ testing and report arithmetic means of the
  per-fold metrics. RMSE is reported on each dataset's MOS scale. Baseline scores
  are taken from the original papers, except for the MM-PCQA and GMS-3DQA results
  on BASICS and LS-PCQA, which are from the retraining reported in
  CLIP-PCQA~\cite{liu2025clippcqa}. \Cref{tab:intra_full} provides the complete
  comparison and baseline protocol details. \textbf{Bold} marks the best score
  and \underline{underlining} marks the second-best score among the values shown
  in each column.}
  \label{tab:intra}
  \small
  \setlength{\tabcolsep}{4pt}
  \resizebox{\textwidth}{!}{\begin{tabular}{l ccc ccc ccc ccc}
    \toprule
    & \multicolumn{3}{c}{SJTU-PCQA}
    & \multicolumn{3}{c}{WPC}
    & \multicolumn{3}{c}{BASICS}
    & \multicolumn{3}{c}{LS-PCQA Part I} \\
    \cmidrule(lr){2-4}\cmidrule(lr){5-7}\cmidrule(lr){8-10}\cmidrule(lr){11-13}
    Method & SRCC & PLCC & RMSE & SRCC & PLCC & RMSE
           & SRCC & PLCC & RMSE & SRCC & PLCC & RMSE \\
    \midrule
    \multicolumn{13}{l}{\textit{Full-reference}} \\
    PCQM~\cite{meynet2020pcqm} & 0.864 & 0.885 & 1.086 & 0.743 & 0.750 & 15.16 & 0.808 & 0.891 & 0.478 & 0.426 & 0.208 & 0.789 \\
    TCDM~\cite{zhang2024tcdm} & 0.930 & 0.910 & 0.891 & 0.807 & 0.804 & 13.53 & 0.757 & 0.874 & 0.505 & 0.413 & 0.438 & 0.739 \\
    \midrule
    \multicolumn{13}{l}{\textit{No-reference: representative methods compared in DSP-PCQA / PMP-PCQA}} \\
    MM-PCQA~\cite{zhang2023mmpcqa} & 0.910 & 0.923 & 0.772 & 0.841 & 0.856 & 12.35 & 0.738 & 0.793 & 0.628 & 0.605 & 0.644 & 0.621 \\
    GMS-3DQA~\cite{zhang2024gms3dqa} & 0.911 & 0.918 & 0.787 & 0.831 & 0.834 & 12.23 & 0.807 & 0.895 & 0.472 & 0.645 & 0.666 & 0.606 \\
    CoPA~\cite{shan2024copa} & 0.897 & 0.913 & 0.739 & 0.779 & 0.785 & 11.93 & -- & -- & -- & -- & -- & -- \\
    CLIP-PCQA~\cite{liu2025clippcqa} & 0.936 & 0.956 & 0.693 & -- & -- & -- & 0.872 & 0.932 & 0.382 & 0.736 & 0.755 & 0.533 \\
    LMM-PCQA~\cite{zhang2024lmmpcqa} & 0.938 & 0.940 & 0.718 & 0.883 & 0.874 & 11.82 & -- & -- & -- & -- & -- & -- \\
    \midrule
    \multicolumn{13}{l}{\textit{No-reference: recent methods}} \\
    Hallucinated-PQA~\cite{mu2024hallucinated} & 0.933 & 0.961 & 0.649 & \underline{0.910} & \underline{0.911} & \underline{9.33} & -- & -- & -- & -- & -- & -- \\
    Dynamic Alignment~\cite{wu2026dynamic} & 0.931 & 0.944 & 0.788 & 0.898 & 0.906 & 10.05 & -- & -- & -- & 0.737 & 0.745 & 0.556 \\
    Com-PCQA~\cite{su2026compcqa} & 0.940 & 0.956 & 0.69 & 0.853 & 0.857 & 11.67 & \textbf{0.907} & \underline{0.940} & 0.399 & 0.745 & 0.733 & 0.590 \\
    DSP-PCQA~\cite{li2026dsp} & 0.951 & 0.964 & 0.637 & 0.892 & 0.889 & 11.57 & -- & -- & -- & -- & -- & -- \\
    SAF-Net~\cite{xiong2026safnet} & 0.949 & 0.963 & -- & 0.898 & 0.896 & -- & -- & -- & -- & -- & -- & -- \\
    R3-PCQA~\cite{seo2026r3pcqa} & 0.940 & 0.961 & 0.723 & 0.902 & 0.888 & 10.96 & -- & -- & -- & -- & -- & -- \\
    PMP-PCQA~\cite{li2026pmp} & \textbf{0.957} & \underline{0.967} & \underline{0.624} & -- & -- & -- & 0.892 & 0.936 & \textbf{0.337} & \underline{0.763} & \underline{0.776} & \underline{0.512} \\
    \midrule
    \textbf{PGS-PCQA (ours)} & \underline{0.955} & \textbf{0.969} & \textbf{0.57} & \textbf{0.934} & \textbf{0.934} & \textbf{8.16} & \underline{0.906} & \textbf{0.946} & \underline{0.340} & \textbf{0.791} & \textbf{0.802} & \textbf{0.487} \\
    \bottomrule
  \end{tabular}}
\end{table*}

On LS-PCQA Part I, using the same PGS-PCQA weights, SRCC is 0.779 with
center-crop testing and 0.791 with foreground-weighted $4\times4$ testing
(\cref{tab:crop}). The intra-dataset splits hold out references, while
training and test sets largely share distortion types. We next evaluate
transfer across datasets.

\noindent\textbf{Cross-dataset.}
Trained on SJTU-PCQA, PGS-PCQA achieves SRCCs of 0.709 on WPC and
0.751 on BASICS, the highest scores shown for these two transfers
in \cref{tab:cross}.

\begin{table}[t]
  \centering
  \caption{Cross-dataset SRCC under the protocol of DSP-PCQA~\cite{li2026dsp}
  (train on the full source, test on the full target). \textbf{Bold}: best,
  \underline{underline}: second; * reproduced in DSP-PCQA, others from the
  original papers.}
  \label{tab:cross}
  \small
  \setlength{\tabcolsep}{3pt}
  \resizebox{\ifdim\width>\linewidth\linewidth\else\width\fi}{!}{\begin{tabular}{lcccc}
    \toprule
    & \multicolumn{2}{c}{SJTU-PCQA$\to$}
    & \multicolumn{2}{c}{WPC$\to$} \\
    \cmidrule(lr){2-3}\cmidrule(lr){4-5}
    Method & WPC & BASICS & SJTU & BASICS \\
    \midrule
    MM-PCQA*~\cite{zhang2023mmpcqa} & 0.201 & 0.378 & 0.755 & 0.600 \\
    GMS-3DQA*~\cite{zhang2024gms3dqa} & 0.396 & 0.268 & 0.684 & 0.469 \\
    CLIP-PCQA*~\cite{liu2025clippcqa} & 0.324 & 0.470 & -- & -- \\
    LMM-PCQA*~\cite{zhang2024lmmpcqa} & 0.475 & 0.652 & 0.783 & 0.626 \\
    M3-Unity~\cite{zhou2024m3unity} & 0.444 & 0.537 & 0.821 & 0.617 \\
    Stochasticity-aware~\cite{fan2025stochasticity} & -- & -- & \textbf{0.913} & -- \\
    Dynamic Alignment~\cite{wu2026dynamic} & 0.340 & -- & 0.801 & -- \\
    DA-Net~\cite{wu2025danet} & 0.401 & -- & 0.743 & -- \\
    Hallucinated-PQA~\cite{mu2024hallucinated} & 0.432 & -- & 0.628 & -- \\
    Com-PCQA~\cite{su2026compcqa} & -- & -- & \underline{0.851} & -- \\
    DSP-PCQA~\cite{li2026dsp} & \underline{0.595} & \underline{0.723} & 0.835 & \underline{0.665} \\
    \midrule
    PGS-PCQA & \textbf{0.709} & \textbf{0.751} & 0.799 & \textbf{0.667} \\
    \bottomrule
  \end{tabular}}
\end{table}

\Cref{tab:cross_pmp} provides results for additional source--target
pairs.

\subsection{Ablation: Synthetic Bank and Perceptual Gate}

The synthetic supervision of PGS-PCQA has two parts: the bank provides level pairs, and the perceptual gate sets the ranking target of each pair. Starting from the model without synthetic supervision (same network and testing as PGS-PCQA), we add hard ranking~\cite{liu2017rankiq,su2022prlgqa} (i.e., a fixed target $p=1$) and then the gate, and also test cross-type ranking. All configurations are trained on SJTU-PCQA.

\begin{table}[t]
  \centering
    \caption{Ablation results. (a) Cross-dataset SRCC for variants of synthetic supervision and network components trained on SJTU-PCQA. (b) SRCC with center-crop and foreground-weighted $4\times4$ testing, using the same weights within each row.}
  \label{tab:ablation}
  \small
  {\footnotesize (a)}\\
  \resizebox{\linewidth}{!}{\begin{tabular}{lccc}
    \toprule
    Configuration & WPC & BASICS & LS-PCQA \\
    \midrule
    w/o PGS & 0.273 & 0.500 & 0.258 \\
    Bank + hard ranking & 0.604 & 0.696 & 0.408 \\
    Bank + perceptual gate (PGS-PCQA) & 0.709 & 0.751 & 0.438 \\
    + cross-type ranking & 0.609 & 0.714 & 0.404 \\
    \midrule
    PGS-PCQA w/o KPA & 0.697 & 0.748 & 0.420 \\
    PGS-PCQA w/o aux. head & 0.709 & 0.729 & 0.418 \\
    \bottomrule
  \end{tabular}}

  \vspace{3pt}
  {\footnotesize (b)}\\
  \resizebox{\linewidth}{!}{\begin{tabular}{lcc}
    \toprule
    Model, setting & Center crop & $4\times4$ test \\
    \midrule
    PGS-PCQA, WPC (intra) & 0.894 & 0.934 \\
    PGS-PCQA, SJTU$\to$WPC & 0.677 & 0.709 \\
    Base network, LS-PCQA (intra) & 0.746 & 0.767 \\
    Com-PCQA, WPC (intra) & 0.815 & 0.848 \\
    \bottomrule
  \end{tabular}}
\end{table}

In \cref{tab:ablation}(a), the hard-ranking configuration has higher
reported SRCC than the model without PGS on all three target
datasets. The perceptual-gate configuration has higher reported
SRCC than hard ranking. In \cref{tab:mech}, dipIQ-style confidence
weighting~\cite{ma2017dipiq} and shuffled soft targets have lower
reported WPC scores than the perceptual gate. The MM-PCQA
comparison also shows higher SRCC with PGS on both target datasets
(\cref{tab:mm_pgs}).

The cross-type ranking configuration has lower reported SRCC than
the within-type configuration on all three target datasets
(\cref{tab:ablation}(a)). We use within-type ranking in the full
model. The configurations without KPA or without the auxiliary
head show different score changes across target datasets.

\subsection{Effect of Multi-Region Testing}

\cref{tab:ablation}(b) fixes the weights and changes only the test cropping. Foreground-weighted $4\times4$ testing helps in every setting and also on another architecture, Com-PCQA~\cite{su2026compcqa}; as a control, averaging 16 repeated center crops of Com-PCQA, whose forward pass samples randomly, gains only 0.003, so the gain comes from wider spatial coverage rather than more predictions. The cross-dataset gain does not depend on it: with the center crop only, SJTU$\to$WPC still reaches 0.677, well above the previous best of 0.595.

\subsection{Generalization across Distortion Types}

Overall SRCC does not show which distortion types fail. We compute SRCC within each target distortion type, comparing the model without synthetic supervision with PGS-PCQA, both trained on SJTU-PCQA (\cref{tab:pertype}).

\begin{table}[t]
  \centering
  \caption{Within-type SRCC for models trained on SJTU-PCQA. Samples are pooled within each distortion type.}
  \label{tab:pertype}
  \small
  \resizebox{\linewidth}{!}{\begin{tabular}{lccc}
    \toprule
    Target distortion type & w/o PGS & PGS-PCQA & In source \\
    \midrule
    WPC: downsampling & 0.840 & 0.879 & yes \\
    WPC: Gaussian noise & 0.823 & 0.833 & yes \\
    WPC: G-PCC Octree & 0.362 & 0.781 & partly (SJTU octree) \\
    WPC: V-PCC & 0.194 & 0.477 & no \\
    WPC: G-PCC Trisoup & $-$0.175 & 0.718 & no \\
    BASICS: V-PCC & 0.306 & 0.781 & no \\
    BASICS: GeoCNN & 0.422 & 0.484 & no \\
    \bottomrule
  \end{tabular}}
\end{table}

The base model still ranks downsampling and Gaussian noise well; both also occur in the source dataset. It is much weaker on coding distortions, and on G-PCC Trisoup its SRCC is $-0.175$, i.e., the predicted order is reversed relative to the subjective scores. %
The reversal also occurs within individual objects. Across runs, 14--16 of the 20 WPC references have negative SRCC across Trisoup levels, indicating that differences in score scales across objects
alone cannot explain the reversal.
The original CLIP-PCQA (retrained on the full SJTU-PCQA with the official settings) reaches only 0.026--0.059 on Trisoup ($4\times4$ and center-crop testing), i.e., chance-level ranking. The reason is that the noise and downsampling of SJTU-PCQA roughen surfaces, so the model learns to prefer smoother ones, while heavier Trisoup approximates the surface with larger triangles and renders smoother with fewer details, and is judged better. Cross-dataset failure is thus concentrated on distortion types absent from the source rather than spread evenly. With synthetic supervision, SRCC rises from $-0.175$ to 0.718 on G-PCC Trisoup, from 0.362 to 0.781 on G-PCC Octree, and from 0.306 to 0.781 on BASICS V-PCC, while downsampling and Gaussian noise barely change (0.840 to 0.879, 0.823 to 0.833). The gain of PGS-PCQA comes mainly from the distortion types missing from the source.

\subsection{Generalization Beyond Synthetic Distortion Coverage}

We examine whether the synthetic bank helps on distortions whose
categories are excluded from that bank. For BASICS, we remove all
G-PCC and V-PCC distortions. For WPC, we remove all codec distortions,
downsampling, and color, brightness, and geometric Gaussian noise.
\Cref{tab:lto} reports the results for the reduced bank, the full
bank, and training without the bank.

\begin{table}[t]
  \centering
  \caption{Cross-dataset SRCC after excluding the corresponding target distortion categories from the synthetic bank.}
  \label{tab:lto}
  \small
  \resizebox{\linewidth}{!}{\begin{tabular}{lcc}
    \toprule
    Configuration & SJTU$\to$WPC & SJTU$\to$BASICS \\
    \midrule
    w/o PGS & 0.273 & 0.500 \\
    Bank w/o target types & 0.555 & 0.645 \\
    Full bank & 0.709 & 0.751 \\
    \bottomrule
  \end{tabular}}
\end{table}

On both WPC and BASICS, the reported SRCC for the reduced bank falls
between those for the full bank and training without the bank.
Relative to training without the bank, the reported reduced-bank
scores represent 65\% and 58\% of the full-bank SRCC gain on WPC
and BASICS, respectively (\cref{tab:lto}). The within-type results
also show higher SRCC for Trisoup and V-PCC with the reduced bank
than without PGS (\cref{tab:lto_type}).

\subsection{Cross-Content Quality Calibration}\label{sec:calib}

To examine transfer to SJTU-PCQA, we compare ranking within
individual objects with the consistency of mean scores across
objects. \Cref{tab:calib} reports mean within-object SRCC and
between-object SRCC for this calibration analysis.

With WPC as the source, the within-object SRCC is 0.900 and the between-object SRCC is 0.433. As a diagnostic, aligning each object's mean predicted score to its mean MOS raises the overall SRCC from
0.800 to 0.886 (\cref{tab:calib}). The model orders the distortion levels of one object well, but its score scale is inconsistent across objects. Synthetic ranking is built between levels of the same reference, so it constrains exactly the within-object order and nothing across objects, and cannot fix this error.

\noindent\textbf{Limitations}
are threefold: the bank helps little on the learning-based GeoCNN codec (0.422 to 0.484, \cref{tab:pertype}); transfers to SJTU-PCQA are limited by cross-object calibration; and foreground-weighted $4\times4$ testing increases inference cost.

\section{Conclusion}
\label{sec:conclusion}

We address the cross-dataset generalization of NR-PCQA limited by insufficient distortion coverage and propose PGS-PCQA, an NR-PCQA model trained with perceptually gated synthetic supervision. It builds a synthetic distortion bank from source references only, uses within-type level relations as extra supervision, and modulates the ranking targets with perceptual distances on rendered views; the synthetic supervision is used only during training and needs no target-domain data. PGS-PCQA substantially improves cross-dataset transfer dominated by uncovered coding distortions and generalizes to distortion types the bank does not cover: even with no coding distortion in the bank, the number of objects with a reversed Trisoup order drops from 14--16 of 20 to 1--5. Since this supervision mainly constrains within-content relations, cross-content score calibration is left for future work. Code and models will be made publicly available.

\section*{LLM Usage Statement}
We used an AI assistant (Claude, Anthropic) for language editing and translation of the manuscript, for writing plotting scripts and preparing figure assets, and for assisting with experiment scripts, including the reproduction of baseline methods.

{
    \small
    \bibliographystyle{ieeenat_fullname}
    \bibliography{refs}
}

\clearpage
\maketitlesupplementary
\appendix
\renewcommand{\topfraction}{0.95}\renewcommand{\bottomfraction}{0.9}\renewcommand{\textfraction}{0.05}\renewcommand{\floatpagefraction}{0.85}\renewcommand{\dbltopfraction}{0.95}\renewcommand{\dblfloatpagefraction}{0.85}\setcounter{dbltopnumber}{2}\setcounter{topnumber}{4}\setcounter{bottomnumber}{3}\setcounter{totalnumber}{6}

\cref{sec:supp_9} first gives more qualitative examples. Besides perceptually gated synthetic supervision, we explored three auxiliary designs to assess geometric feature supplementation, distortion-attribute supervision, and test-time spatial coverage. The base predictor is CLIP-PCQA~\cite{liu2025clippcqa}. Their implementation and results are given in \cref{sec:supp_1,sec:supp_2,sec:supp_3}; \cref{sec:supp_4} gives the full intra-dataset comparison, \cref{sec:supp_5} details of the synthetic bank, and \cref{sec:supp_7} the PMP-PCQA source--target pairs, \cref{sec:supp_mech,sec:supp_ltotype,sec:supp_mm} the comparison with related ranking mechanisms, the per-type results of the exclusion experiment, and PGS on MM-PCQA, and \cref{sec:supp_8,sec:supp_10} perceptual-gate statistics and cross-content calibration. Reference numbers follow the main paper.

\section{More Qualitative Examples}\label{sec:supp_9}

\begin{figure*}[t]
  \centering
  \includegraphics[width=\textwidth]{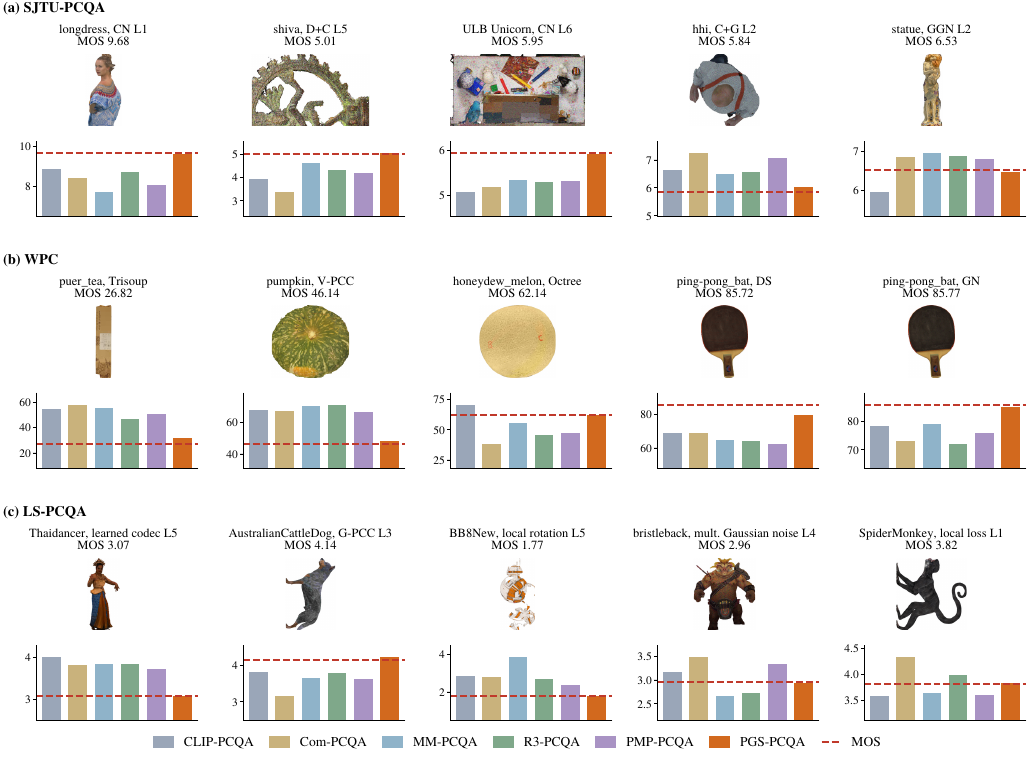}
  \caption{Additional qualitative comparisons on SJTU-PCQA, WPC, and LS-PCQA Part I. Bars show
  held-out predictions mapped to the MOS scale by a per-fold logistic fit. The red
  dashed line shows the subjective MOS. The methods reproduced for this figure use the
  same reference folds as in \cref{fig:qual}. On SJTU-PCQA, CN denotes color noise,
  D+C denotes downscaling plus color noise, GGN denotes geometry Gaussian noise, and
  C+G denotes color plus geometry Gaussian noise. L indicates distortion level. On
  WPC, Trisoup and Octree are G-PCC geometry codecs, DS denotes downsampling, and GN
  denotes Gaussian noise. WPC MOS ranges from 0 to 100. On LS-PCQA Part I, the learned
  codec denotes learning-based coding, and MOS ranges from 1 to 5. Across all
  evaluated samples, the mean absolute errors of PGS-PCQA are 0.48 on SJTU-PCQA, 6.5
  on WPC, and 0.38 on LS-PCQA Part I. The corresponding ranges for the compared
  methods are 0.51--0.62, 7.4--10.1, and 0.39--0.46.}
  \label{fig:qual_full}
\end{figure*}

\cref{fig:qual_full} gives more qualitative comparisons with other methods on SJTU-PCQA, WPC, and LS-PCQA.

\section{Auxiliary Modules}\label{sec:supp_1}

\noindent\textbf{KPA}
is a geometry branch operating directly on 3D points, inspired by local-neighborhood aggregation for point clouds~\cite{wu2019pointconv,thomas2019kpconv}; it supplements point density and local hole information that 2D projections lose. It samples 2048 query points and aggregates neighborhood features at three scales ($h = 0.003$, $0.006$, $0.012$) with at most 32 neighbors each, combining Gaussian distance weights, learnable gates, and a density correction. The fused features pass through two Transformer layers and pooling into a 768-d representation, which corrects the CLIP quality logits through a zero-initialized linear layer. An extra MOS regression loss with weight 0.5 is used during training.

\noindent\textbf{Distortion auxiliary head}
adds distortion-type classification and parameter regression on the shared visual features, following multi-task learning with distortion attributes~\cite{shan2024gpanet}. Labels are parsed from file names, the loss weight is 0.5, and the head is removed at inference.

\noindent\textbf{Foreground-weighted $4\times4$ testing}
reduces the information loss of center cropping: we take $4\times4$ uniformly placed $224\times224$ windows on each projected view, predict quality for each of the 16 cross-view window groups, and fuse them weighted by their foreground-pixel ratio averaged over six views. This changes no weights but increases inference cost. All main experiments use this protocol.

\section{Module Ablation}\label{sec:supp_2}

\Cref{tab:abl_ls} compares combinations of the auxiliary modules and PGS on LS-PCQA
Part I. PGS refers to the synthetic bank with perceptually gated within-type
ranking. Each configuration uses one training run per fold in the same five-fold
reference-disjoint split, the epoch-50 checkpoint, and foreground-weighted
$4\times4$ testing.

\begin{table}[H]
  \centering
  \caption{Intra-dataset ablation of the auxiliary modules and PGS on LS-PCQA Part I. All
  configurations use the same five reference-disjoint folds and the evaluation
  procedure described in \cref{sec:supp_2}.}
  \label{tab:abl_ls}
  \small
  \resizebox{\linewidth}{!}{\begin{tabular}{cccccc}
    \toprule
    KPA & Aux. head & PGS & SRCC$\uparrow$ & PLCC$\uparrow$ & RMSE$\downarrow$ \\
    \midrule
    \ding{51} & -- & -- & 0.762 & 0.778 & 0.513 \\
    -- & \ding{51} & -- & 0.778 & 0.798 & 0.493 \\
    -- & -- & \ding{51} & 0.786 & 0.797 & 0.493 \\
    \ding{51} & \ding{51} & -- & 0.767 & 0.788 & 0.503 \\
    \ding{51} & -- & \ding{51} & 0.773 & 0.794 & 0.495 \\
    -- & \ding{51} & \ding{51} & 0.792 & 0.805 & 0.484 \\
    \ding{51} & \ding{51} & \ding{51} & 0.791 & 0.802 & 0.487 \\
    \bottomrule
  \end{tabular}}
\end{table}

In the matched LS-PCQA configurations, adding PGS increases SRCC with the auxiliary
head alone, with KPA alone, and with both modules. \Cref{tab:ablation}(a) reports the cross-dataset comparisons for models trained on SJTU-PCQA.
The effects of KPA and the auxiliary head vary across target datasets.

\section{Multi-Region Testing}\label{sec:supp_3}

\Cref{tab:crop} compares center-crop and foreground-weighted $4\times4$ testing
using identical weights within each row. For PGS-PCQA on LS-PCQA Part I, both
settings use the checkpoints at epoch 50 from the five folds in \cref{tab:intra}. In
the Com-PCQA control in \cref{tab:ablation}(b), averaging 16 passes of the same
center crop increases SRCC by 0.003. This control measures the effect of repeated
stochastic predictions on the same crop.

\begin{table}[H]
  \centering
  \caption{SRCC with center-crop and foreground-weighted $4\times4$ testing. The two columns
  use identical weights within each row. The base network is trained without the
  synthetic distortion bank.}
  \label{tab:crop}
  \small
  \resizebox{\linewidth}{!}{\begin{tabular}{lcc}
    \toprule
    Model, dataset & Center crop & $4\times4$ test \\
    \midrule
    Base network (same weights), LS-PCQA & 0.746 & 0.767 \\
    PGS-PCQA, LS-PCQA & 0.779 & 0.791 \\
    Com-PCQA~\cite{su2026compcqa}, WPC & 0.815 & 0.848 \\
    \bottomrule
  \end{tabular}}
\end{table}

\section{Full Intra-Dataset Comparison}\label{sec:supp_4}

\Cref{tab:intra_full} extends the comparison in \cref{tab:intra} with additional
full-reference and no-reference methods. The PGS-PCQA row reports the same results
as \cref{tab:intra}. The table notes identify baseline scores obtained under
different or unspecified evaluation protocols.

\begin{table*}[!t]
  \centering
  \caption{Full intra-dataset quality prediction results (SRCC$\uparrow$, PLCC$\uparrow$,
  RMSE$\downarrow$). PGS-PCQA follows the protocol in \cref{tab:intra}, with the
  epoch-50 checkpoint in each reference-disjoint fold. The comparison includes methods
  from DSP-PCQA~\cite{li2026dsp} and PMP-PCQA~\cite{li2026pmp} and additional recent
  no-reference methods. Scores for classical methods follow the DSP-PCQA and PMP-PCQA
  summaries, which draw on CLIP-PCQA retraining for BASICS and LS-PCQA. Other scores
  follow the original papers unless marked. $^{b}$ marks scores from the DSP-PCQA or
  PMP-PCQA summaries that were unreported in the original papers and whose data splits
  are unspecified. A dash indicates unavailable results. $^{\dagger}$ DA-Net uses
  LS-PCQA Part III with pseudo MOS and a random 8:1:1 split. $^{\ddagger}$ DSP-PCQA
  uses a single 8:1:1 split on BASICS. $^{\S}$ M3-Unity uses a single 60/20/20 split
  on BASICS. $^{\P}$ SAF-Net leaves its LS-PCQA subset unspecified.
  \textbf{Bold} marks the best score and \underline{underlining} marks the second-best score among eligible entries in each column. Entries marked $^{b}$, $^{\dagger}$, $^{\ddagger}$, $^{\S}$, or $^{\P}$ use different or unspecified protocols and are excluded from ranking.}
  \label{tab:intra_full}
  \small
  \setlength{\tabcolsep}{4pt}
  \resizebox{\textwidth}{!}{\begin{tabular}{l ccc ccc ccc ccc}
    \toprule
    & \multicolumn{3}{c}{SJTU-PCQA} & \multicolumn{3}{c}{WPC} & \multicolumn{3}{c}{BASICS} & \multicolumn{3}{c}{LS-PCQA Part I} \\
    \cmidrule(lr){2-4}\cmidrule(lr){5-7}\cmidrule(lr){8-10}\cmidrule(lr){11-13}
    Method & SRCC & PLCC & RMSE & SRCC & PLCC & RMSE & SRCC & PLCC & RMSE & SRCC & PLCC & RMSE \\
    \midrule
    \multicolumn{13}{l}{\textit{Full-reference}} \\
    MSE-p2po~\cite{mekuria2017codec} & 0.729 & 0.812 & 1.361 & 0.456 & 0.485 & 19.89 & 0.774 & 0.849 & 0.559 & 0.301 & 0.427 & 0.744 \\
    MSE-p2pl~\cite{tian2017geometric} & 0.628 & 0.594 & 2.282 & 0.328 & 0.270 & 22.82 & 0.836 & 0.895 & 0.468 & 0.286 & 0.454 & 0.734 \\
    PCQM~\cite{meynet2020pcqm} & 0.864 & 0.885 & 1.086 & 0.743 & 0.750 & 15.16 & 0.808 & 0.891 & 0.478 & 0.426 & 0.208 & 0.789 \\
    GraphSIM~\cite{yang2022graphsim} & 0.878 & 0.845 & 1.032 & 0.583 & 0.616 & 17.19 & 0.813 & 0.895 & 0.465 & 0.331 & 0.358 & 0.767 \\
    TCDM~\cite{zhang2024tcdm} & 0.930 & 0.910 & 0.891 & 0.807 & 0.804 & 13.53 & 0.757 & 0.874 & 0.505 & 0.413 & 0.438 & 0.739 \\
    \midrule
    \multicolumn{13}{l}{\textit{No-reference: methods compared in DSP-PCQA / PMP-PCQA}} \\
    IT-PCQA~\cite{yang2022itpcqa} & 0.865 & 0.828 & 1.166 & 0.487 & 0.433 & 19.90 & 0.310$^{b}$ & 0.302$^{b}$ & 1.024$^{b}$ & -- & -- & -- \\
    3D-NSS~\cite{zhang2022nss} & 0.714 & 0.738 & 1.769 & 0.648 & 0.651 & 16.57 & 0.617$^{b}$ & 0.657$^{b}$ & 0.883$^{b}$ & 0.532 & 0.584 & 0.682 \\
    ResSCNN~\cite{liu2023lspcqa} & 0.880 & 0.889 & 0.878 & 0.435 & 0.429 & 23.27 & 0.352 & 0.391 & 0.975 & 0.620 & 0.648 & 0.615 \\
    MM-PCQA~\cite{zhang2023mmpcqa} & 0.910 & 0.923 & 0.772 & 0.841 & 0.856 & 12.35 & 0.738 & 0.793 & 0.628 & 0.605 & 0.644 & 0.621 \\
    GMS-3DQA~\cite{zhang2024gms3dqa} & 0.911 & 0.918 & 0.787 & 0.831 & 0.834 & 12.23 & 0.807 & 0.895 & 0.472 & 0.645 & 0.666 & 0.606 \\
    PIT-PCQA~\cite{xie2024pitpcqa} & 0.903 & 0.927 & 0.725 & -- & -- & -- & 0.831$^{b}$ & 0.851$^{b}$ & 0.514$^{b}$ & 0.712$^{b}$ & 0.723$^{b}$ & 0.574$^{b}$ \\
    CoPA~\cite{shan2024copa} & 0.897 & 0.913 & 0.739 & 0.779 & 0.785 & 11.93 & 0.785$^{b}$ & 0.812$^{b}$ & 0.534$^{b}$ & 0.613$^{b}$ & 0.636$^{b}$ & 0.594$^{b}$ \\
    CLIP-PCQA~\cite{liu2025clippcqa} & 0.936 & 0.956 & 0.693 & -- & -- & -- & 0.872 & 0.932 & 0.382 & 0.736 & 0.755 & 0.533 \\
    LMM-PCQA~\cite{zhang2024lmmpcqa} & 0.938 & 0.940 & 0.718 & 0.883 & 0.874 & 11.82 & 0.845$^{b}$ & 0.867$^{b}$ & 0.443$^{b}$ & 0.731$^{b}$ & 0.744$^{b}$ & 0.547$^{b}$ \\
    \midrule
    \multicolumn{13}{l}{\textit{No-reference: recent methods}} \\
    M3-Unity~\cite{zhou2024m3unity} & 0.947 & 0.961 & -- & 0.900 & 0.900 & -- & 0.872$^{\S}$ & 0.937$^{\S}$ & --$^{\S}$ & -- & -- & -- \\
    Stochasticity-aware~\cite{fan2025stochasticity} & 0.947 & 0.964 & -- & 0.874 & 0.877 & -- & -- & -- & -- & -- & -- & -- \\
    DQP-PCQA~\cite{duan2025dqp} & -- & -- & -- & 0.816 & 0.817 & 12.78 & 0.874 & 0.924 & 0.426 & -- & -- & -- \\
    Hallucinated-PQA~\cite{mu2024hallucinated} & 0.933 & 0.961 & 0.649 & \underline{0.910} & \underline{0.911} & \underline{9.33} & -- & -- & -- & -- & -- & -- \\
    DA-Net~\cite{wu2025danet} & 0.922 & 0.926 & -- & 0.893 & 0.892 & -- & -- & -- & -- & 0.795$^{\dagger}$ & 0.800$^{\dagger}$ & --$^{\dagger}$ \\
    Dynamic Alignment~\cite{wu2026dynamic} & 0.931 & 0.944 & 0.788 & 0.898 & 0.906 & 10.05 & -- & -- & -- & 0.737 & 0.745 & 0.556 \\
    EGMS-PCQM~\cite{liu2026egms} & 0.941 & 0.963 & -- & 0.877 & 0.871 & -- & -- & -- & -- & -- & -- & -- \\
    Com-PCQA~\cite{su2026compcqa} & 0.940 & 0.956 & 0.69 & 0.853 & 0.857 & 11.67 & \textbf{0.907} & \underline{0.940} & 0.399 & 0.745 & 0.733 & 0.590 \\
    DSP-PCQA~\cite{li2026dsp} & 0.951 & 0.964 & 0.637 & 0.892 & 0.889 & 11.57 & 0.875$^{\ddagger}$ & 0.946$^{\ddagger}$ & 0.347$^{\ddagger}$ & -- & -- & -- \\
    SAF-Net~\cite{xiong2026safnet} & 0.949 & 0.963 & -- & 0.898 & 0.896 & -- & -- & -- & -- & 0.884$^{\P}$ & 0.880$^{\P}$ & --$^{\P}$ \\
    R3-PCQA~\cite{seo2026r3pcqa} & 0.940 & 0.961 & 0.723 & 0.902 & 0.888 & 10.96 & -- & -- & -- & -- & -- & -- \\
    PMP-PCQA~\cite{li2026pmp} & \textbf{0.957} & \underline{0.967} & \underline{0.624} & -- & -- & -- & 0.892 & 0.936 & \textbf{0.337} & \underline{0.763} & \underline{0.776} & \underline{0.512} \\
    \midrule
    \textbf{PGS-PCQA (ours)} & \underline{0.955} & \textbf{0.969} & \textbf{0.57} & \textbf{0.934} & \textbf{0.934} & \textbf{8.16} & \underline{0.906} & \textbf{0.946} & \underline{0.340} & \textbf{0.791} & \textbf{0.802} & \textbf{0.487} \\
    \bottomrule
  \end{tabular}}
\end{table*}

\section{Synthetic Distortion Bank}\label{sec:supp_5}

For every source reference point cloud, coordinates are first quantized to a 10-bit voxel grid, and 84 distorted versions are generated (\cref{tab:bank}). The distortion types follow the Moving Picture Experts Group (MPEG) common test conditions and common distortion categories, fixed in advance without regard to any target dataset. Synthetic samples are rendered exactly as real ones. The perceptual distance $d(\cdot)$ is a learned perceptual image patch similarity (LPIPS)-style~\cite{zhang2018lpips} deep-feature distance without learned layer weights: we take the relu1\_2, relu2\_2, relu3\_4, relu4\_4, and relu5\_4 features of an ImageNet-pretrained VGG19~\cite{simonyan2015vgg}, unit-normalize them along channels, average the squared difference between the distorted and reference renders over space, and average the five layers uniformly; renders are resized to at most 512 pixels on the long side, and the result is averaged over six views. It is computed once for all synthetic samples before training, and VGG19 is not trained.

\begin{table}[H]
  \centering
  \caption{Composition of the synthetic distortion bank (per reference point cloud).}
  \label{tab:bank}
  \small
  \begin{tabular}{p{0.2\linewidth}p{0.58\linewidth}c}
    \toprule
    Category & Content & Versions \\
    \midrule
    G-PCC (TMC13 v23) & octree + RAHT, octree + lifting, color only (lossless geometry + RAHT), 5 levels each; trisoup, 4 levels & 19 \\
    V-PCC (TMC2 v25) & all-intra, r1--r5 & 5 \\
    Generic & color Gaussian / salt-and-pepper / luminance noise, contrast / saturation / brightness shift, colour bit-depth quantization, local missing / shift / rotation, geometric Gaussian displacement, random downsampling; 5 levels each & 60 \\
    \bottomrule
  \end{tabular}
\end{table}

\section{Source--Target Pairs of PMP-PCQA}\label{sec:supp_7}

\begin{table}[H]
  \centering
  \caption{Source--target pairs used by PMP-PCQA~\cite{li2026pmp} (SRCC). * reproduced in PMP-PCQA; ``--'': not listed. References shared with SJTU-PCQA are removed from LS-PCQA. $^{\dagger}$ Removal of shared references not specified; excluded from ranking. In addition, PGS-PCQA reaches 0.795 on LS-PCQA$\to$WPC and 0.590 on WPC$\to$LS-PCQA, versus 0.504 and 0.601 for Dynamic Alignment~\cite{wu2026dynamic}.}
  \label{tab:cross_pmp}
  \small
  \resizebox{\ifdim\width>\linewidth\linewidth\else\width\fi}{!}{\begin{tabular}{lcccc}
    \toprule
    & \multicolumn{2}{c}{SJTU-PCQA$\to$} & \multicolumn{2}{c}{LS-PCQA$\to$} \\
    \cmidrule(lr){2-3}\cmidrule(lr){4-5}
    Method & LS & BASICS & SJTU & BASICS \\
    \midrule
    MM-PCQA*~\cite{zhang2023mmpcqa} & -- & 0.178 & 0.723 & 0.423 \\
    GMS-3DQA*~\cite{zhang2024gms3dqa} & -- & 0.568 & 0.776 & 0.459 \\
    CLIP-PCQA*~\cite{liu2025clippcqa} & -- & 0.625 & \underline{0.812} & 0.496 \\
    LMM-PCQA*~\cite{zhang2024lmmpcqa} & 0.405 & 0.552 & 0.662 & 0.511 \\
    M3-Unity~\cite{zhou2024m3unity} & -- & 0.537 & -- & -- \\
    Dynamic Alignment~\cite{wu2026dynamic} & 0.368$^{\dagger}$ & -- & 0.868$^{\dagger}$ & -- \\
    PMP-PCQA~\cite{li2026pmp} & \underline{0.431} & \underline{0.678} & \textbf{0.843} & \underline{0.560} \\
    \midrule
    PGS-PCQA & \textbf{0.438} & \textbf{0.751} & 0.804 & \textbf{0.623} \\
    \bottomrule
  \end{tabular}}
\end{table}

\section{Comparison with Related Ranking Mechanisms}\label{sec:supp_mech}

\Cref{tab:mech} compares ranking methods trained on SJTU-PCQA.
Hard ranking follows RankIQA and PRL-GQA~\cite{liu2017rankiq,su2022prlgqa},
and listwise ranking uses the ListNet loss as in
LRL-GQA~\cite{cao2007listnet,li2025lrlgqa}. The dipIQ-style
variant~\cite{ma2017dipiq} uses hard targets and weights each pair
by $2p-1$. The random gate shuffles the soft targets among level
pairs in the bank. We set $\tau_d$ from the standard deviation of
the synthetic perceptual distances. The table reports the default
gate and variants that halve or double $\tau_d$.

\begin{table}[H]
  \centering
  \caption{Cross-dataset SRCC for different ranking supervision methods
trained on SJTU-PCQA.}
  \label{tab:mech}
  \small
  \begin{tabular}{lcc}
    \toprule
    Ranking supervision & WPC & LS-PCQA \\
    \midrule
    Hard ranking (RankIQA / PRL-GQA) & 0.604 & 0.408 \\
    Listwise ranking (LRL-GQA) & 0.586 & 0.387 \\
    dipIQ-style confidence weighting & 0.633 & 0.419 \\
    Random gate & 0.614 & 0.384 \\
    Perceptual gate, $\tau_d$ halved & 0.645 & 0.421 \\
    Perceptual gate, $\tau_d$ doubled & 0.680 & 0.388 \\
    Perceptual gate (PGS-PCQA) & 0.709 & 0.438 \\
    \bottomrule
  \end{tabular}
\end{table}

\section{Per-Type Results of the Exclusion Experiment}\label{sec:supp_ltotype}

\Cref{tab:lto_type} reports within-type SRCC for selected coding
distortions in the exclusion experiment. For Trisoup on WPC and
V-PCC on BASICS, the reduced-bank model has higher within-type
SRCC than the model without PGS. We also report G-PCC Octree for
reference because SJTU-PCQA contains a related Octree distortion.

\begin{table}[H]
  \centering
  \caption{Within-type SRCC after excluding the corresponding target
distortion categories from the synthetic bank. Models are trained on
SJTU-PCQA. Among the 20 WPC references, 14--16 have negative SRCC
across Trisoup levels without PGS, and 1--5 have negative SRCC
after the target categories are removed}
  \label{tab:lto_type}
  \small
  \begin{tabular}{l>{\centering\arraybackslash}p{1.45cm}>{\centering\arraybackslash}p{1.45cm}>{\centering\arraybackslash}p{1.45cm}}
    \toprule
    Target distortion type & w/o PGS & PGS (full bank) & PGS (target types removed) \\
    \midrule
    WPC: G-PCC Trisoup & $-0.175$ & 0.718 & 0.317 \\
    WPC: G-PCC Octree & 0.362 & 0.781 & 0.740 \\
    WPC: V-PCC & 0.194 & 0.477 & 0.269 \\
    BASICS: V-PCC & 0.306 & 0.781 & 0.493 \\
    \bottomrule
  \end{tabular}
\end{table}

\section{PGS on MM-PCQA}\label{sec:supp_mm}

To test whether PGS depends on the CLIP architecture, we add the same bank and the same perceptually gated ranking loss to the official code of MM-PCQA~\cite{zhang2023mmpcqa} (image and point branches, no CLIP), keep its official hyperparameters, train on SJTU-PCQA under the same cross-dataset protocol, and test the epoch with the lowest training loss. \Cref{tab:mm_pgs} shows that adding PGS to MM-PCQA improves cross-dataset SRCC on both target datasets.

\begin{table}[H]
  \centering
  \caption{Cross-dataset SRCC of MM-PCQA with and without PGS.
Both configurations are trained on SJTU-PCQA.}
  \label{tab:mm_pgs}
  \small
  \begin{tabular}{lcc}
    \toprule
    Model & WPC & BASICS \\
    \midrule
    MM-PCQA & 0.211 & 0.417 \\
    MM-PCQA + PGS & \textbf{0.335} & \textbf{0.531} \\
    \bottomrule
  \end{tabular}
\end{table}

\section{Perceptual Gate Statistics}\label{sec:supp_8}

\begin{table}[H]
  \centering
  \caption{Perceptual separability of adjacent levels of the same distortion type in the synthetic banks. The WPC bank contains 19 references (tool\_box is not included), and 10 Trisoup versions failed to encode in the BASICS bank, so the pair counts are below 67 per reference.}
  \label{tab:gate}
  \small
  \begin{tabular}{l>{\centering\arraybackslash}p{1.55cm}>{\centering\arraybackslash}p{1.55cm}>{\centering\arraybackslash}p{1.55cm}}
    \toprule
    Bank built from & Adjacent pairs & Perceptual order reversed & Soft target below 0.75 \\
    \midrule
    SJTU-PCQA & 603 & 4.1\% & 23.4\% \\
    WPC & 1273 & 4.9\% & 22.2\% \\
    BASICS & 5015 & 3.5\% & 21.8\% \\
    LS-PCQA & 5695 & 4.1\% & 22.9\% \\
    \bottomrule
  \end{tabular}
\end{table}

\begin{table}[H]
  \centering
\caption{Agreement between the perceptual gate and real subjective scores (level pairs of the same reference and distortion type). Agreement is the share of pairs whose MOS follows the preset level order. $\rho(\Delta d)$ and $\rho(\text{level diff.})$ are the Spearman correlations of the perceptual-distance difference and of the level difference with $\Delta$MOS; pairs are split into quintiles by $\Delta d$, and Agr.\ (low/high) is the agreement for the 20\% of pairs with the smallest/largest $\Delta d$. Adjacent: pairs of adjacent levels only.}
  \label{tab:gatemos}
  \footnotesize
  \setlength{\tabcolsep}{3pt}
  \resizebox{\linewidth}{!}{%
  \begin{tabular}{lcccccc}
    \toprule
    Dataset & Level pairs & Pairs & $\rho(\Delta d)$ & $\rho(\text{level diff.})$ & Agr.\ (low) & Agr.\ (high) \\
    \midrule
    SJTU-PCQA & all & 945 & 0.797 & 0.664 & 0.799 & 1.000 \\
     & adjacent & 315 & 0.457 & -- & 0.714 & 0.937 \\
    WPC & all & 2109 & 0.775 & 0.643 & 0.865 & 1.000 \\
     & adjacent & 874 & 0.535 & -- & 0.829 & 0.994 \\
    BASICS & all & 3064 & 0.852 & 0.588 & 0.865 & 1.000 \\
     & adjacent & 1195 & 0.734 & -- & 0.774 & 1.000 \\
    \bottomrule
  \end{tabular}%
  }
\end{table}

\section{Cross-Content Quality Calibration}\label{sec:supp_10}

\cref{tab:calib} gives the full results of \cref{sec:calib}.

\begin{table}[H]
  \centering
  \caption{Cross-content calibration with SJTU-PCQA as the target (SRCC).}
  \label{tab:calib}
  \small
  \setlength{\tabcolsep}{3pt}
  \resizebox{\linewidth}{!}{\begin{tabular}{lcccc}
    \toprule
    Source & Overall & Within-object & Between-object & MOS-aligned (upper bound) \\
    \midrule
    WPC & 0.800 & 0.900 & 0.433 & 0.886 \\
    LS-PCQA & 0.784 & 0.838 & 0.206 & 0.859 \\
    \bottomrule
  \end{tabular}}
\end{table}

\end{document}